\documentclass{article}

\usepackage[preprint]{neurips_2026}

\usepackage[utf8]{inputenc} % allow utf-8 input
\usepackage[T1]{fontenc}    % use 8-bit T1 fonts
\usepackage{hyperref}       % hyperlinks
\usepackage{url}            % simple URL typesetting
\usepackage{booktabs}       % professional-quality tables
\usepackage{amsfonts}       % blackboard math symbols
\usepackage{nicefrac}       % compact symbols for 1/2, etc.
\usepackage{microtype}      % microtypography
\usepackage{xcolor}         % colors
\usepackage{graphicx}
\usepackage{amsmath}
\usepackage{subcaption}
\usepackage{caption}
\usepackage{tabularx}
\usepackage{float}
\usepackage{longtable}
\usepackage{wrapfig}
\usepackage{multirow}
\usepackage{lmodern}

\title{OceanCBM: A Concept Bottleneck Model for Mechanistic Interpretability in Ocean Forecasting}

\author{%
  Sanah Suri\thanks{Corresponding author: sansuri@ucdavis.edu} \\
  Department of Computer Science\\
  University of California, Davis\\
  Davis, CA 95616 \\
  \And
  Kieran Ringel \\
  Department of Computer Science\\
  University of California, Davis\\
  Davis, CA 95616 \\
  \And
  Maike Sonnewald\textsuperscript{1,2} \\
  Department of Computer Science\\
  University of California, Davis\\
  Davis, CA 95616 \\
}

\begin{document}

\maketitle
\footnotetext[1]{University of Washington, Seattle, WA, USA.}
\footnotetext[2]{NOAA Geophysical Fluid Dynamics Laboratory, Princeton, NJ, USA.}

\begin{abstract}
  Extreme ocean phenomena are challenging not only to predict but to diagnose, as accurate forecasts alone do not reveal the underlying physical drivers. While recent machine learning approaches achieve strong predictive skill, they remain largely opaque and provide limited guarantees of fidelity to ground-truth physics. We introduce OceanCBM, the first concept bottleneck model (CBM) for spatiotemporal prediction and mechanistic interrogation of ocean dynamics. OceanCBM uses mixed supervision to predict mixed layer heat content, a key precursor of marine heatwaves, while routing information through an intermediate layer of prescribed concepts derived from geophysical fluid dynamics and a 'free' concept. This design imposes soft physical structure without over-constraining the model, and the free concept both regularizes concept predictions and captures residual physical processes. Across ensemble initializations, we show that mixed supervision yields consistent mechanistic representations, whereas prediction-only and prescription-only baselines learn highly variable latent structures despite similar predictive performance. OceanCBM achieves interpretable, physically grounded representations without sacrificing skill, explicitly characterizing the interpretability-performance trade-off.
\end{abstract}

\section{Introduction}\label{s:intro}
Machine learning (ML) has transformed our ability to not just predict future states of  physical systems, but also learn more about them \cite{cheng2023machine}. However, as deeper and more complex neural networks (NNs) trained on high-dimensional data become more prevalent, their reasoning becomes harder to translate into human-interpretable meaning \cite{hassija2024interpreting}. As interest in ML interpretability grows, driven in part by the widespread deployment of large language models, a key insight has emerged: understanding why a model makes a prediction can be as valuable as the prediction itself \cite{singh2024rethinking}. This is especially acute in high-stakes scientific domains, where a model that predicts correctly for the wrong reasons may fail precisely when reliability matters most: under novel conditions, during extreme events, or in a changing climate. Our proposed methodology addresses this directly, balancing skill with trust. 

ML approaches have shown considerable promise in emulating the Earth system, forecasting ocean and atmospheric states with remarkable skill \cite{lang2024aifs, lam2023learning, duncan2025samudrace}. However, a skillful model may have learned spurious correlations rather than meaningful underlying physics. In light of this, research has moved towards embedding known physical laws into a model's training pipeline, or validating learned representations through post-hoc analyses. Physics-informed neural networks (PINNs) enforce dynamical constraints directly into the model's learning, offering physical fidelity \cite{karniadakis2021physics, raissi2019physics}. However, these constraints can be too rigid to generalize to regimes outside the historical record and the problem of possible convergence to physically incorrect spurious solutions persists \cite{krishnapriyan2021characterizing}. Post-hoc approaches such as explainable artificial intelligence (XAI) methods interrogate a trained model to detect which inputs it relied upon \cite{flora2024machine}. However, this analysis explains what the model did rather than ensuring the physics was adhered to \cite{suri2026trusting}.  

Thus, existing approaches either enforce structure or infer structure after the fact, offering no guarantee that the learned mechanisms are physically meaningful. Concept bottleneck models (CBMs) offer a middle ground by routing predictions through an intermediate layer of interpretable concepts. They structurally embed physical reasoning into the model without over-constraining it. We present OceanCBM, a novel adaptation of this framework to ocean physics, designed to recover the mechanistic drivers of marine heatwaves (MHWs) through prescribed concepts and a free concept that captures residual structure and regularizes prescription. By combining mixed supervision with an ensemble of CBMs, OceanCBM balances physical constraint, flexibility, and mechanistic interpretability.

MHWs are characterized as periods of prolonged anomalously high sea surface temperatures (SSTs) \cite{hobday2016hierarchical}. They are growing in frequency and intensity, with socioeconomic impacts ranging from marine ecosystem collapse to hurricane intensification, affecting communities worldwide \cite{spillman2025makes}. Critically, MHW impacts are likely to compound with other stressors in ways that remain poorly characterized, making mechanistic understanding urgent \cite{capotondi2024global}. While MHWs are defined by SST anomalies, the processes that drive them are not confined to the surface, but occur in the mixed layer, the interface where the ocean interacts with the atmosphere. For example, horizontal and vertical transport of heat by ocean currents and mixing dynamics have been found to play a critical role in their formation and persistence \cite{sun2023frequent, ren2026ocean}. Understanding these mechanistic drivers therefore requires turning to an integrated measure of upper ocean heat, namely the mixed layer heat content (MLHC) \cite{vijith2020closing}. Thus, instead of forecasting SST, we target MLHC and related subsurface drivers of ocean heat, enabling a physically grounded, mechanistic interpretation of MHW formation.

\begin{figure}[h]
    \centering
    \includegraphics[width=\linewidth]{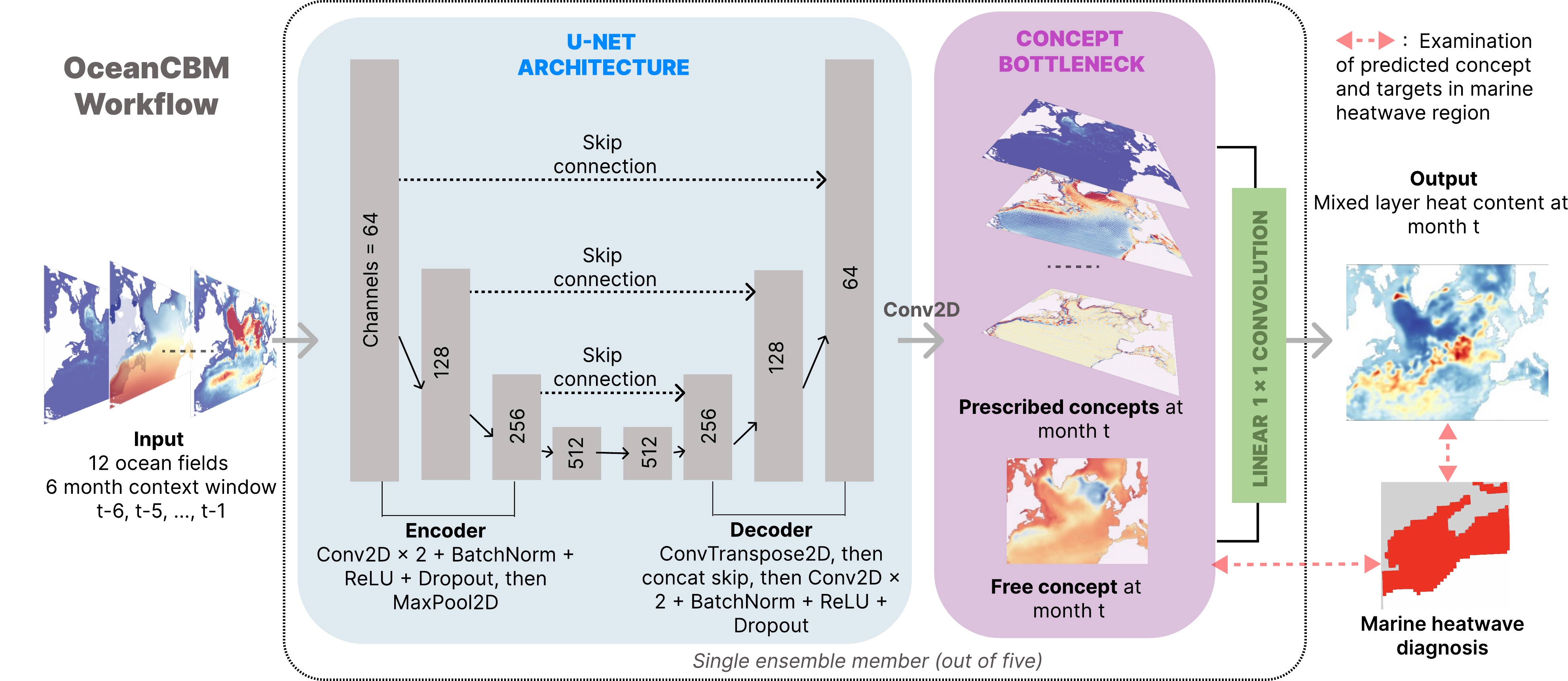}
    \caption{\textbf{Workflow diagram}. OceanCBM is an NN ensemble that utilizes a U-Net architecture to predict mixed layer heat content, using prescribed concepts and a free concept. The predicted fields are used to study regions with recorded marine heatwaves.}
    \label{fig:architecture}
\end{figure}

Our contributions are:
\begin{itemize}
    \item We introduce OceanCBM, the \textbf{first concept bottleneck framework for spatiotemporal ocean physics prediction}, enabling skillful forecasting of MLHC and interpretable, physically grounded interrogation into the drivers of MHWs. 
    \item We propose a \textbf{mixed supervision bottleneck} combining prescribed physical concepts with a free concept that serves a dual role: \textbf{regularizing concept predictions} and providing a \textbf{structured channel for emergent phenomena} not captured by the prescribed set.
    \item We show that mixed supervision \textbf{reduces variation in learned mechanistic pathways} across random initializations in the ensemble, whereas prediction-only and prescription-only baselines discover inconsistent latent representations despite comparable predictive skill.
    \item We apply OceanCBM to the 2012 Gulf of Maine MHW, demonstrating recovery of known and new physical drivers in a \textbf{real-world setting}.
\end{itemize}

\section{Related Work}
\paragraph{Concept bottleneck models.} Concept bottleneck models (CBMs) are an interpretable ML architecture in which models predict human-interpretable concepts as intermediate representations before making a final prediction. Rather than directly learning the target function $f(x)$, the model learns a function composition $f = g\circ h (x)$, where $h$ predicts intermediate concepts and $g$ maps these concepts to the target. The function $g$ is typically linear, making the final prediction step interpretable by design. Since their formalization in \cite{koh2020cbm} and successful application to both classification and regression tasks on image data, CBMs have been applied to several domains to impart interpretability. In \cite{sawada2022concept}, a CBM is trained to leverage both supervised and unsupervised concepts to learn image data while in \cite{oikarinen2023label}, the authors eliminate the need for labeled concept data entirely by leveraging foundation models to automatically generate concept sets. Separately, \cite{van2024timeseries} applied a concept bottleneck within the attention mechanism of a transformer for long-term time series forecasting. However, these approaches are domain-agnostic and lack the physical grounding that scientific applications require. As such, \cite{asiyabi2026pgcbm} applied process-guided CBMs to biomass estimation using physically motivated concepts, but prescribed a fixed concept set with no residual channel for unrepresented processes and did not address spatiotemporal prediction. Additional limitations of CBM implementations include a performance trade-off for interpretability and a non-exhaustive concept set \cite{margeloiu2021leakage, yuksekgonul2022post, oikarinen2023label}. We address these limitations by demonstrating strong predictive skill and including a free concept to capture residual information.

\paragraph{Physics guided machine learning.} ML models have an established propensity to extract meaningful information from data and can outperform traditional statistical and numerical methods, motivating their use for atmospheric and oceanic forecasting \cite{sonnewald2021bridging, irrgang2021towards}. Data-driven methods such as \cite{dheeshjith2025samudra, duncan2025samudrace, guo2025data, lam2023learning, willard2022integrating}, achieve remarkable accuracy and are trained on state-of-the-art datasets, but cannot be trusted for operational forecasting due to their black-box nature. As such, interpretable and explainable approaches have emerged, where model learning is guided by ground-truth physics or validated against it \cite{suri2026trusting, yik2023southern, sonnewald2021revealing}. The authors of \cite{rudin2022interpretable} advocate building inherently interpretable models, especially in domains where model prediction can inform high-stakes decision making.  Efforts in this direction range from fidelity verification \cite{suri2026trusting}, post-hoc explainability analyses \cite{flora2024machine, yik2023southern} and causal methods that aim to identify mechanistic drivers \cite{camps2025review, iglesias2024causally}. Our work lies at the confluence of such methods in that we use a mixed supervision approach to gently guide the model to known physics, while leaving room for unknown phenomena and using an ensemble to reveal varying mechanistic learning pathways.  

\section{Methods}\label{s:methods}

\paragraph{ORAS5 dataset.} Reanalysis products are foundational to data-driven forecasting in the Earth system, providing long-term records that combine model output with observational constraints \cite{hersbach2020era5}. While ERA5, provided by the European Center for Medium-Range Weather Forecasts (ECMWF), has become the standard for atmospheric ML applications, its oceanic counterpart presents untapped opportunities. We use the Ocean Reanalysis System 5 (ORAS5), which not only provides a long historical ocean record but also comprises $5$ ensemble members, giving an estimate of internal variability that single-member reanalysis products do not offer. Each member represents a physically plausible trajectory of the ocean state, capturing uncertainty from observational and atmospheric forcing errors. ORAS5 consists of monthly averaged ocean fields at $1/4^\circ$ horizontal resolution spanning the years 1979--2018 \cite{zuo2019ecmwf}. To the best of our knowledge, OceanCBM is the first model to leverage all five ORAS5 ensemble members as training data, treating each member-month as an independent physically consistent sample. Other applications typically use the control member or ensemble mean \cite{pasula2025data, guo2025data}. The ORAS5 ensemble members capture data uncertainty in the ocean state, while the OceanCBM ML ensemble members capture model uncertainty arising from random weight initialization.

\paragraph{Input.} Since we predict MLHC, we consider variables from ORAS5 at relevant points of the mixed layer, collapsed to 2D spatial fields: some are taken directly at the surface, others are evaluated at the base of the mixed layer, and others are averaged across the mixed layer to represent integrated subsurface dynamics. Surface fields such as sea surface height capture atmospheric interactions, while depth-evaluated fields such as zonal and meridional ocean velocity (the east-west and north-south components of ocean currents) represent subsurface dynamics.

\begin{figure*}[t]
\centering

\begin{minipage}[c]{0.59\textwidth}
    \centering
    \includegraphics[width=\linewidth]{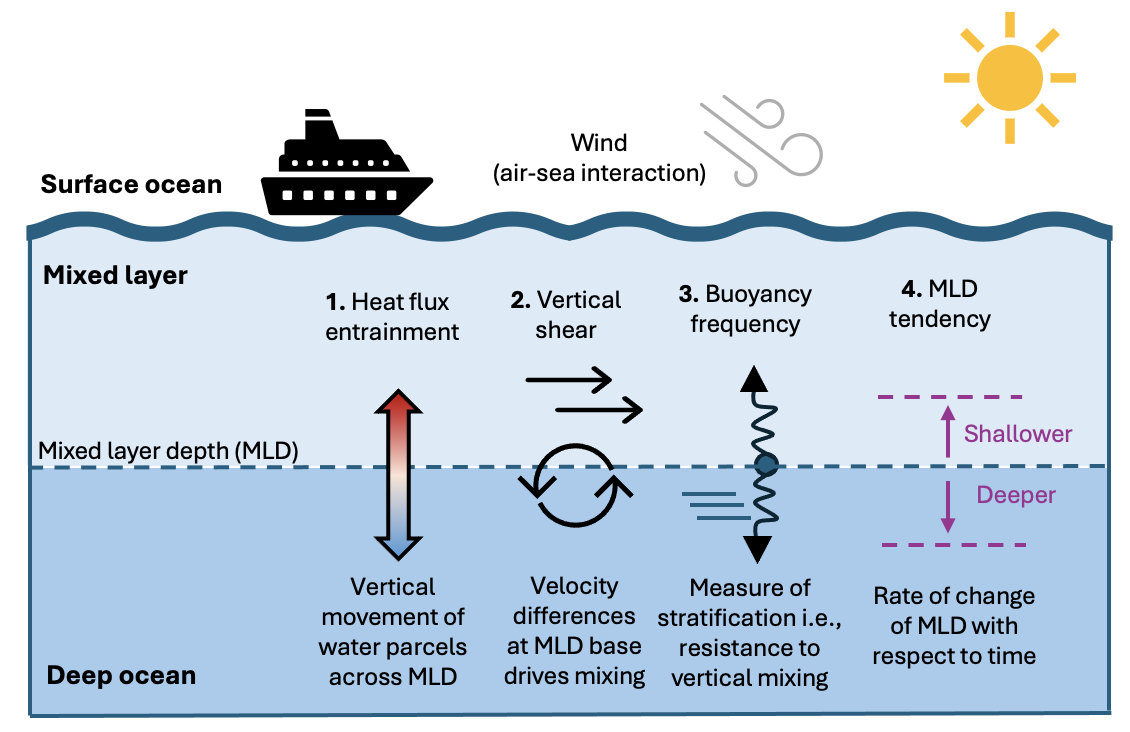}
\end{minipage}
\hfill
\begin{minipage}[c]{0.4\textwidth}
    \centering
    \includegraphics[width=\linewidth]{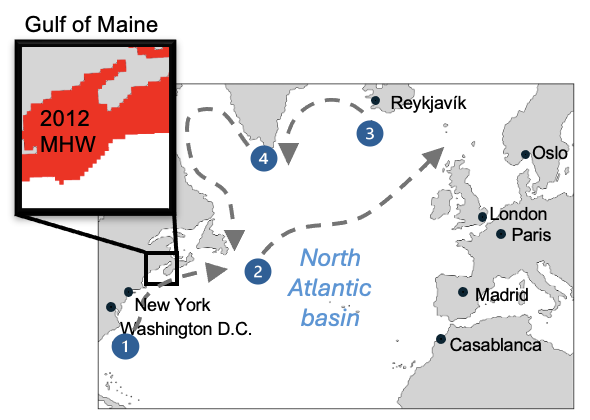}
\end{minipage}

\caption{\textbf{Concepts and geographical region.}
Left panel: The prescribed concepts and their role in impacting the mixed layer heat content.
Right: North Atlantic geographic region in consideration, the domain of the 2012 Gulf of Maine marine heatwave, used as a case study and the key ocean currents in the region: (1) Gulf Stream, (2) North Atlantic drift, (3) Subpolar gyre, and (4) Labrador current.}
\label{fig:concept_na}
\end{figure*}

\paragraph{Prescribed concepts.} The mixed layer is the upper part of the ocean where dynamics are challenging to resolve because of the interacting physics of the ocean and the atmosphere, and can directly impact MHW formation \cite{vijith2020closing}. The prescribed concepts in the left panel of Fig. \ref{fig:concept_na} are guided by well-established physical processes controlling MLHC. These include (1) heat flux entrainment (heat exchange across the mixed layer), (2) vertical shear (velocity differences that promote mixing), (3) buoyancy frequency (how strongly the water resists mixing due to density differences), and (4) mixed layer depth (MLD) tendency (rate of deepening or shoaling of the mixed layer). These quantities are computed from ORAS5 input fields and evaluated at the surface, at the MLD base, or averaged across the mixed layer. We note that while the concepts are chosen to represent some of the main known drivers of mixed layer dynamics, the system is not fully observable since MLHC is challenging to measure directly, and additional mechanisms may be undiscovered or unresolved in data, motivating the addition of the free concept. 

\paragraph{Preprocessing.} All ocean fields are cropped to the North Atlantic basin, corresponding to the right panel of Fig. \ref{fig:concept_na}. This region includes major circulation features and contains a historically significant MHW, the 2012 Gulf of Maine event, which we analyze as a case study later on. Each input instance is a $6$ month context window of $12$ 2D physical variables on a $302\times 400$ spatial grid, labeled with target concepts and target MLHC, predicted at a $1$ month lead time. All inputs and prescribed concepts are detrended by fitting and subtracting a linear trend over the $40$ year dataset, such that the model learns relevant climatological variability rather than a warming trend. We split the data temporally to prevent leakage, using an $80-10-10$ training-validation-test split.

\paragraph{Model architecture.} OceanCBM is an ensemble of $5$ U-Net based CBMs \cite{ronneberger2015u}. The encoder consists of $4$ downsampling stages with channel widths $[64, 128, 256, 512]$, with each stage applying $3\times 3$ convolutions, with batch normalization and ReLU activation, followed by $2\times 2$ max pooling. At the deepest level, $2$ more convolutions are applied. The decoder mirrors the encoder with $4$ upsampling operations using transposed convolutions. It outputs a supervised concept head, producing $4$ prescribed concept fields and a free concept head producing one unconstrained field. These are then passed through a linear layer that combines the channels to predict the target MLHC at $302\times 400$ spatial resolution, $1$ month in the future. Fig. \ref{fig:architecture} describes the workflow in detail.
\paragraph{Training.} We train OceanCBM $5$ times with independent random weight initializations to assess robustness to initialization, optimizing a weighted combination of concept prediction loss and MLHC prediction loss (both $\ell_1$):
\begin{equation}
\label{eq:loss}
    \mathcal{L}_{\text{combined}} = \lambda \mathcal{L}_{\text{concept}} + (1-\lambda)\mathcal{L}_{\text{pred}}
\end{equation}
We employ an adaptive $\lambda$ schedule that begins at $\lambda = 0.8$ and decays exponentially to $\lambda = 0.2$ over $100$ epochs. This schedule prioritizes concept learning early in training, allowing the bottleneck to establish physically meaningful representations, while ending at $\lambda=0.2$ to maintain a degree of concept supervision throughout.

\paragraph{Model assessment.} We evaluate predictive skill of MLHC and the prescribed concepts using the anomaly correlation coefficient (ACC), a spatial Pearson correlation metric used operationally by ECMWF and National Oceanic and Atmospheric Administration (NOAA) for forecast verification \cite{murphy1989skill}. ACC quantifies pattern similarity between predicted and target fields by first removing the monthly climatology, computed per grid cell over the full evaluation period, from both prediction and target, and then computing the Pearson correlation between the resulting anomalies. This metric captures how well the model reproduces the spatial structure of ocean fields, rather than the mean state or point-wise accuracy. To evaluate the interpretability–prediction trade-off, we consider two baselines: (1) a prediction-only model using the same architecture without the concept bottleneck layer but with matching latent dimensions, and (2) a prescription-only model that retains the concept bottleneck but removes the free concept. These baselines are further used to evaluate consistency in mechanistic learning relative to OceanCBM and to assess the effect of concept regularization. 

\section{Results}\label{s:results}

In this section we discuss OceanCBM's ensemble skill and variability. We demonstrate strong spatial and temporal skill using seasonally averaged spatial ACC maps and basin-averaged time series comparisons between predicted and target fields over the validation and test sets. We focus on the target, mixed layer heat content (MLHC), and two of the four prescribed concepts: buoyancy frequency and heat flux entrainment. We further examine the mechanistic pathways learned across ensemble members and the dual role of the free concept as a residual channel and concept regularizer, with vertical shear serving as an illustrative example.
\begin{figure}[ht]
    \centering
    % --- LEFT COLUMN: Model Accuracy ---
    \begin{subfigure}[b]{0.5\textwidth}
        \centering
        \includegraphics[width=\textwidth]{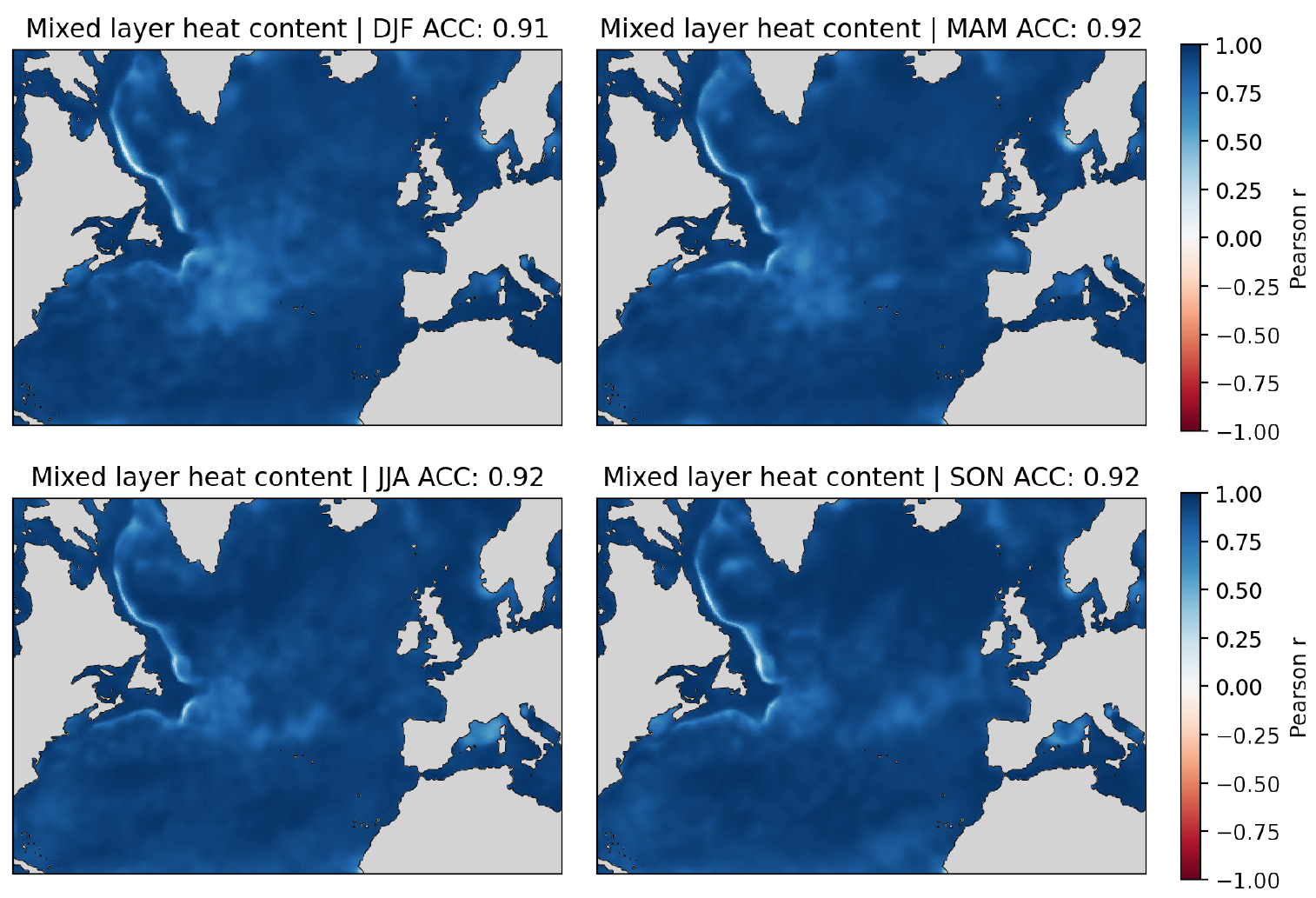}
        \caption{Out-of-sample seasonal ACC average}
    \end{subfigure}
    \hfill
    % --- RIGHT COLUMN: Stacked Time Series and Table ---
    \begin{minipage}[b]{0.45\textwidth}
        \centering
        % Top part: Time Series
        \begin{subfigure}{\textwidth}
            \centering
            \includegraphics[width=\textwidth]{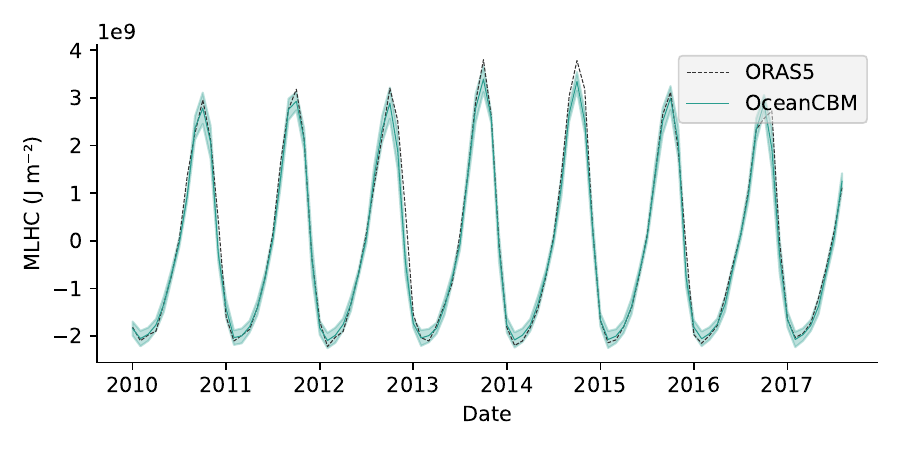}
            \vspace{-1.5em}
            \caption{Basin-averaged time series prediction}
        \end{subfigure}
        
        \vspace{0.5em}
        
        % Bottom part: The Table
        \begin{subfigure}{\textwidth}
            \centering
            \resizebox{\textwidth}{!}{% Scale table to fit column width
            \begin{tabular}{llcccc}
                \toprule
                 \textbf{Model} & \textbf{DJF} & \textbf{MAM} & \textbf{JJA} & \textbf{SON} \\ \midrule
                 OceanCBM & 0.9116 & 0.9174 & 0.9191 & 0.9207 \\
                 Prediction-only & 0.9131 & 0.9186 & 0.9180 & 0.9216 \\
                     Prescription-only & 0.9129 & 0.9187 & 0.9188 & 0.9212 \\
                \bottomrule
            \end{tabular}}
            \caption{Seasonal ACC comparison}
        \end{subfigure}
    \end{minipage}

    \caption{\textbf{OceanCBM skill.} (a) depicts seasonally averaged spatial ACC of MLHC, (b) shows high alignment of basin-averaged time-series prediction with the ground-truth where shading represents model spread ($\pm 2$ standard deviations), and (c) summarizes seasonal ACC for MLHC from baseline models and OceanCBM.}
    \label{fig:mlhc_performance}
\end{figure}

\begin{figure}[h]
    \centering
    % Row 1: VOHFE
    \begin{subfigure}[b]{0.4\textwidth}
        \includegraphics[width=\textwidth]{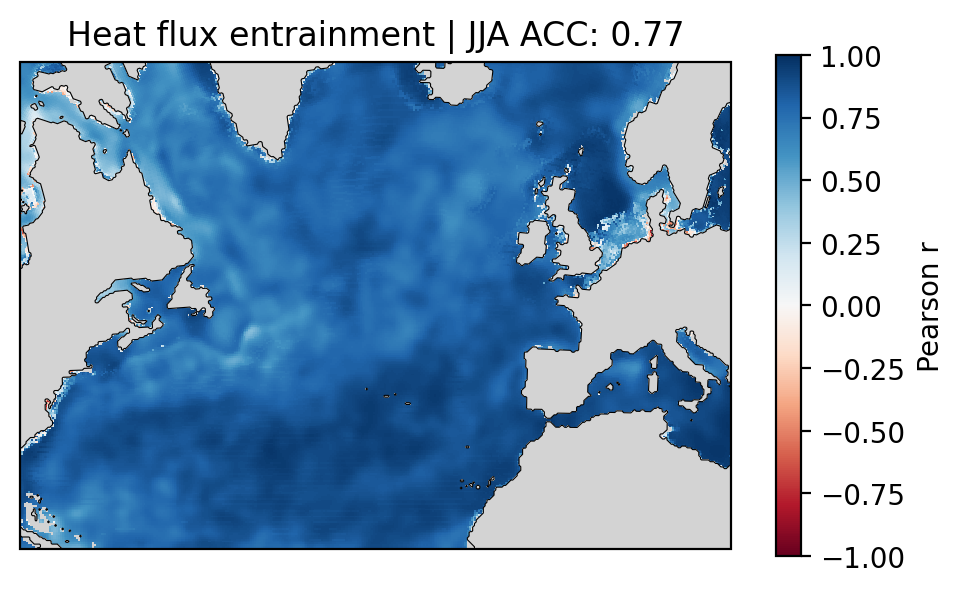}
        \caption{Heat flux entrainment (JJA)}
    \end{subfigure}
    \hfill
    \begin{subfigure}[b]{0.55\textwidth}
        \includegraphics[width=\textwidth]{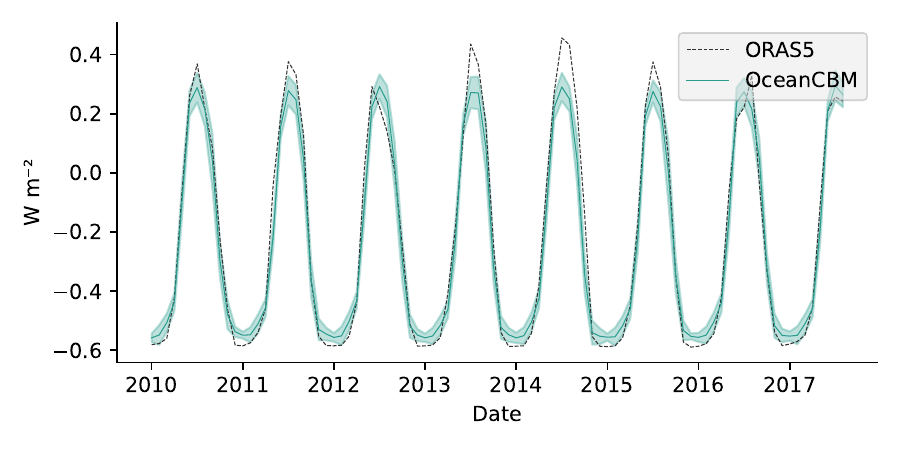}
        \caption{Heat flux entrainment time series}
    \end{subfigure}

    \vspace{0.3cm}

    % Row 3: VON2
    \begin{subfigure}[b]{0.4\textwidth}
        \includegraphics[width=\textwidth]{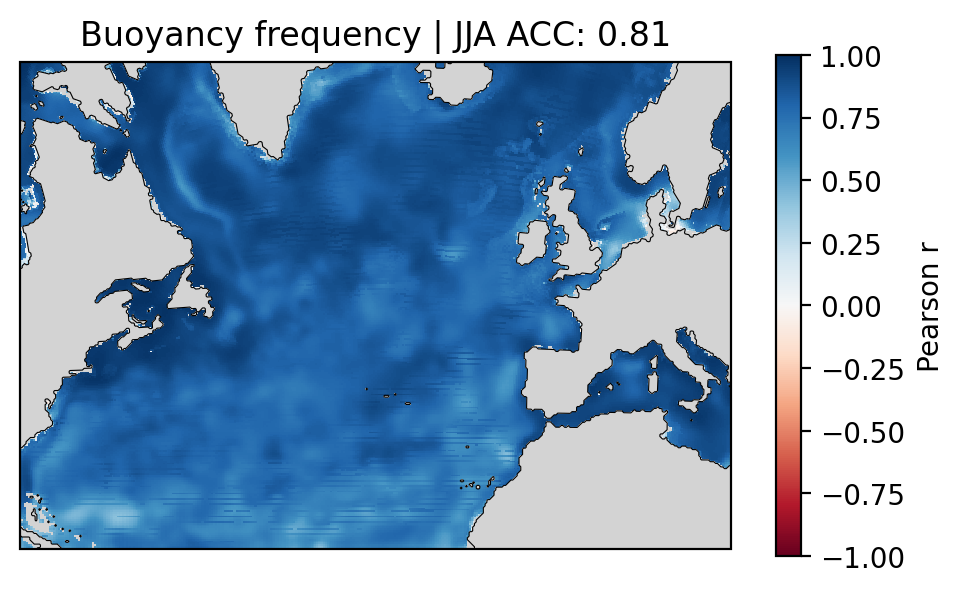}
        \caption{Buoyancy frequency (JJA)}
    \end{subfigure}
    \hfill
    \begin{subfigure}[b]{0.55\textwidth}
        \includegraphics[width=\textwidth]{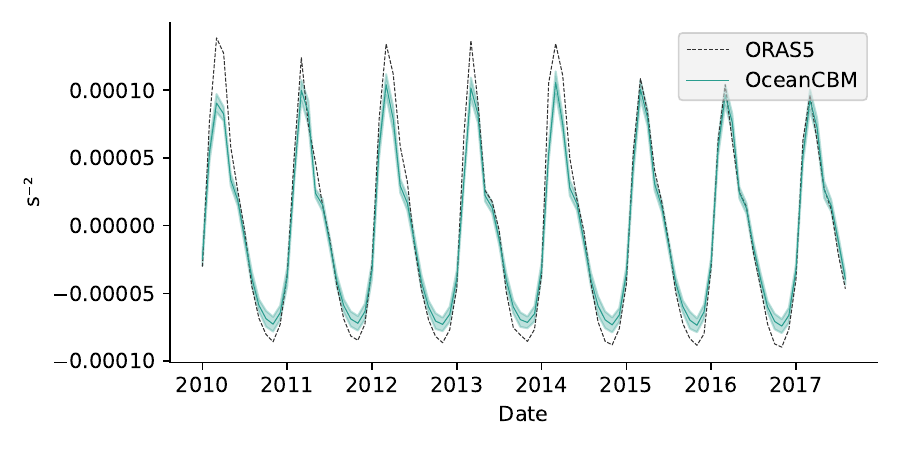}
        \caption{Buoyancy frequency time series}
    \end{subfigure}

    \caption{\textbf{Concept skill.} Left column shows the spatial ACC in predicting (a) heat flux entrainment and (b) buoyancy frequency during summer months (June-July-August). Right column compares the basin-averaged predicted concept values against ORAS5 ground-truth. The shading represents $\pm 2$ standard deviations across the ensemble members, to visualize model spread.}
    \label{fig:concept_mechanisms}
\end{figure}

\paragraph{Spatial and temporal skill.} As an upper bound on predictive skill, we first consider the prediction-only model, which achieves a mean ACC of $0.92$ across seasons in the validation and test sets. In comparison, OceanCBM attains nearly identical performance, with seasonal ACC values ranging from $0.91$ to $0.92$ (Fig.~\ref{fig:mlhc_performance} (a)), indicating essentially no loss in predictive skill from introducing interpretability constraints. This behavior is consistent in the basin-averaged MLHC time series over the validation and test periods (Fig.~\ref{fig:mlhc_performance} (b)), where OceanCBM accurately reproduces the seasonal cycle. The lowest correlations are localized in dynamically active regions, including the Gulf Stream, North Atlantic Drift, and Labrador Current, marked in Fig. \ref{fig:concept_na}. Here, prediction is more challenging due to rapidly changing ocean conditions. However, these reductions in skill are modest, and the ACC field remains physically structured, indicating that the model captures the dominant basin-scale variability.

\paragraph{Concept prediction.} For the concepts, the ACC values overall are lower than the target field. However, this is expected. Unlike MLHC, the concepts are optimized jointly through a shared loss and are learned partly through their contribution to the target prediction. This reflects a deliberate design choice: the concepts are intended to capture physically meaningful mechanisms relevant to prediction, rather than serve as perfect deterministic surrogates for the target. Notably, concept predictions remain skillful.  In Fig. \ref{fig:concept_mechanisms} (a) and (c), we find that in June-July-August (JJA) heat flux entrainment achieves an ACC of $0.77$ and buoyancy frequency has a maximum ACC of $0.81$. These metrics indicate the ensemble is learning meaningful representations of the concepts rather than simply approximating them weakly for MLHC prediction. In Fig. \ref{fig:concept_mechanisms}(b) and (d), we observe strong temporal alignment of predicted and true concepts in the out-of-sample time period.

\begin{figure}[htbp]
     \centering
     \begin{subfigure}[b]{0.31\linewidth}
         \centering
         \includegraphics[width=\linewidth]{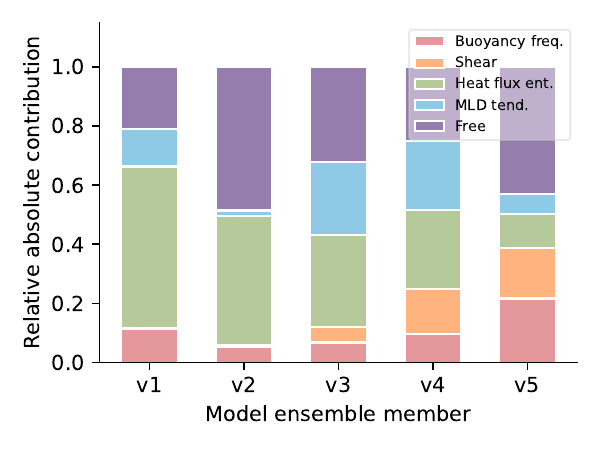}
         \caption{OceanCBM model.}
         \label{fig:weights_concept}
     \end{subfigure}
     \hfill
     \begin{subfigure}[b]{0.31\linewidth}
         \centering
         \includegraphics[width=\linewidth]{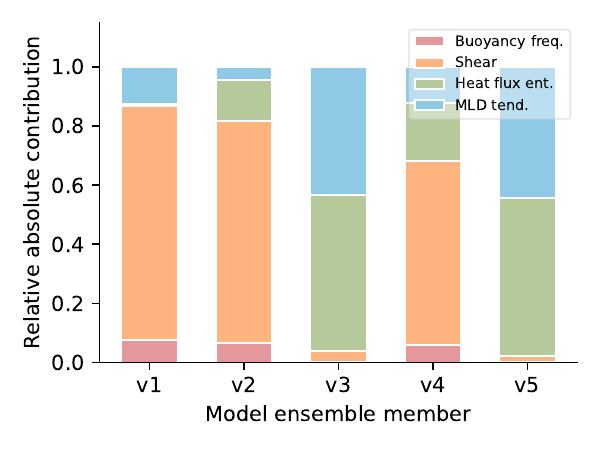}
         \caption{Prescription-only model.}
         \label{fig:weights_pres}
     \end{subfigure}
     \hfill
     \begin{subfigure}[b]{0.31\linewidth}
         \centering
         \includegraphics[width=\linewidth]{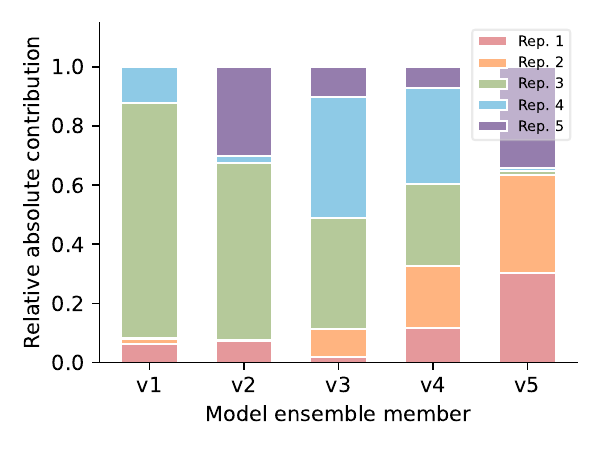}
         \caption{Prediction-only model.}
         \label{fig:weights_latent}
     \end{subfigure}
     \caption{\textbf{Comparing weights of bottleneck.} All models use identical initialization seeds across v1–v5 to ensure controlled comparison. Values are computed by taking absolute value of the weights and normalizing by their sum.}
     \label{fig:weights_comparison}
\end{figure}

\paragraph{Consistency in learned mechanistic pathways.} Fig. \ref{fig:weights_comparison}  respectively highlights (a) the contribution of the prescribed and free concepts in the bottleneck layer of OceanCBM, (b) the contribution of the concepts in the prescription-only baseline, and (c) the contribution of the learned latent representations in the prediction-only baseline. Recall from Fig. \ref{fig:architecture}, the final layer of each model is a linear mapping from concept representations to the target, equivalent to a multilinear regression. We define contribution as the normalized absolute regression weights, yielding relative contributions in between $0$ and $1$ for comparison across model configurations. In these configurations, we note the prediction skill is relatively uniform across ensemble members, but the internal model representations and concept usage are clearly operating differently, indicating multiple viable mechanistic pathways to similar predictions. In Fig. \ref{fig:weights_concept}, across ensemble members of OceanCBM, the weight distribution varies. This suggests two important points: (1) our prescribed concept set is not exhaustive and can be expanded to capture additional physical structure, and (2) ensembling is a necessity. The multiple viable pathways effect across ensemble members is significantly amplified in Fig. \ref{fig:weights_pres} and Fig. \ref{fig:weights_latent}. The prescription-only baseline (Fig. \ref{fig:weights_pres}) primarily seems to either latch onto vertical shear, or a combination of MLD tendency and heat flux entrainment, but does not assign much weight to the rest while the prediction-only baseline (Fig. \ref{fig:weights_latent}) shows variation in representation learning across the ensemble members, making it impossible to discern a clear learning strategy. While Fig. \ref{fig:weights_concept} indicates that OceanCBM is sensitive to random initialization, it exhibits notably more consistency in mechanistic learning than its prediction-only and prescription-only counterparts. This suggests that the mixed supervision approach of prescribed concepts and a free concept is essential for guiding the model toward physically meaningful learning pathways.

\begin{figure}[htbp]
    \centering
    % --- Left Side: The Image ---
    \begin{minipage}{0.56\textwidth}
        \centering
        \includegraphics[width=\linewidth]{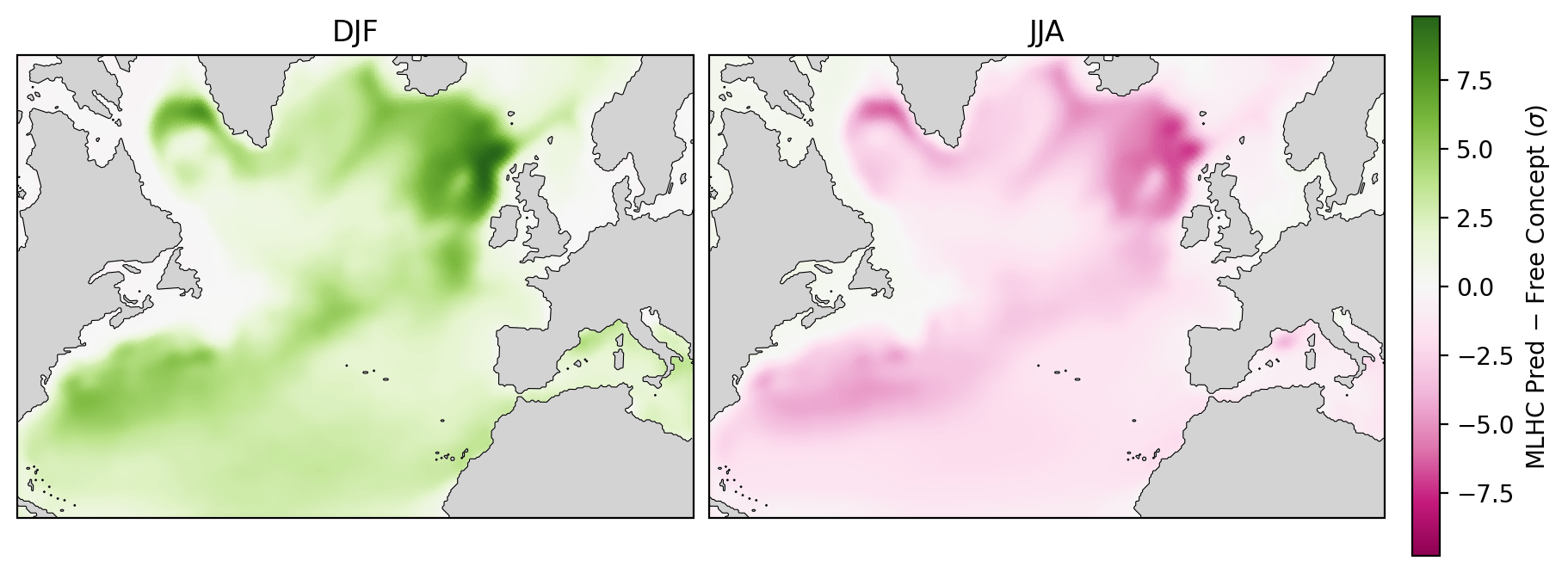}
        \caption{\textbf{Seasonal differences between MLHC and free concept.} The difference between the predicted MLHC and free concept, taken in normalized space, over the out-of-sample time period. Results are shown for DJF and JJA, representing winter and summer conditions respectively.}
        \label{fig:bias}
    \end{minipage}
    \hfill
    % --- Right Side: The Table ---
    \begin{minipage}{0.4\textwidth}
        \centering
        {\small \textbf{Vertical Shear ACC}} \\
        \vspace{0.5em}
        \small
        \begin{tabular}{lcc}
            \toprule
             & OceanCBM & Prescription-only \\
            \midrule
            DJF & \textbf{0.60} & 0.53 \\
            MAM & \textbf{0.61} & 0.57 \\
            JJA & \textbf{0.67} & 0.57 \\
            SON & \textbf{0.58} & 0.57 \\
            \bottomrule
        \end{tabular}
        \captionof{table}{\textbf{Comparing concept ACC:} Seasonal mean ACC of vertical shear for OceanCBM and prescription-only model, with ACC being consistently lower in the prescription-only configuration.}
        \label{tab:s2}
    \end{minipage}
\end{figure}

\paragraph{Free concept analysis.} We turn our attention to the free concept, demonstrating its dual purpose as  a residual channel for emergent physics and a prescribed concept regularizer. The left panel of Fig. \ref{fig:bias} plots the difference between the free concept and predicted MLHC in the winter (December-January-February or DJF) and the summer (June-July-August or JJA). Regions where the difference is near zero indicate that the prescribed concepts and free concept contribute similarly to the prediction. Large positive or negative differences indicate regions where the prescribed concepts and free concept diverge, capturing complementary physical signals. These differences are spatially coherent and concentrated along dynamically active regions, specifically the ocean currents marked in Fig. \ref{fig:concept_na}, and reverse sign seasonally, consistent with changing ocean-atmosphere dynamics between winter and summer. The free concept thus captures residual predictive structure not represented by the prescribed concepts while the seasonal discrepancies noted demonstrate that the prediction is not a trivial reconstruction of the unconstrained free concept. Importantly, viewing Fig. \ref{fig:weights_concept} and Fig. \ref{fig:bias} together demonstrates that dominance in linear weight by the free concept does not imply sole control of the prediction, as the spatial discrepancies show the predicted MLHC is not simply reducible to the free concept, addressing the risk of concept leakage. We observe the concept regularization phenomena empirically in Tab.~\ref{tab:s2}, which compares vertical shear ACC between OceanCBM and the prescription-only baseline. While MLHC predictive skill remains comparable across the two model configurations (Fig.~\ref{fig:mlhc_performance} (c)), performance on vertical shear decreases substantially. Given that vertical shear is among the most challenging concepts to predict, this suggests that the free concept absorbs residual variability that would otherwise be allocated to shear, acting as a structured regularizer within the bottleneck.
\section{Case Study}\label{s:study}
A diagnostic analysis of the $2012$ Gulf of Maine MHW demonstrates the real-world viability of OceanCBM. The right panel of Fig. \ref{fig:concept_na} denotes the Gulf of Maine region and highlights in red exactly the areas where a MHW was observed. To interrogate whether the MLHC and our concepts align with what is known about the Gulf of Maine MHW, we do a $6$ month retrospective of model predictions before the MHW is observed in August 2012. 
\begin{figure}[h]
\centering
\includegraphics[width=\linewidth]{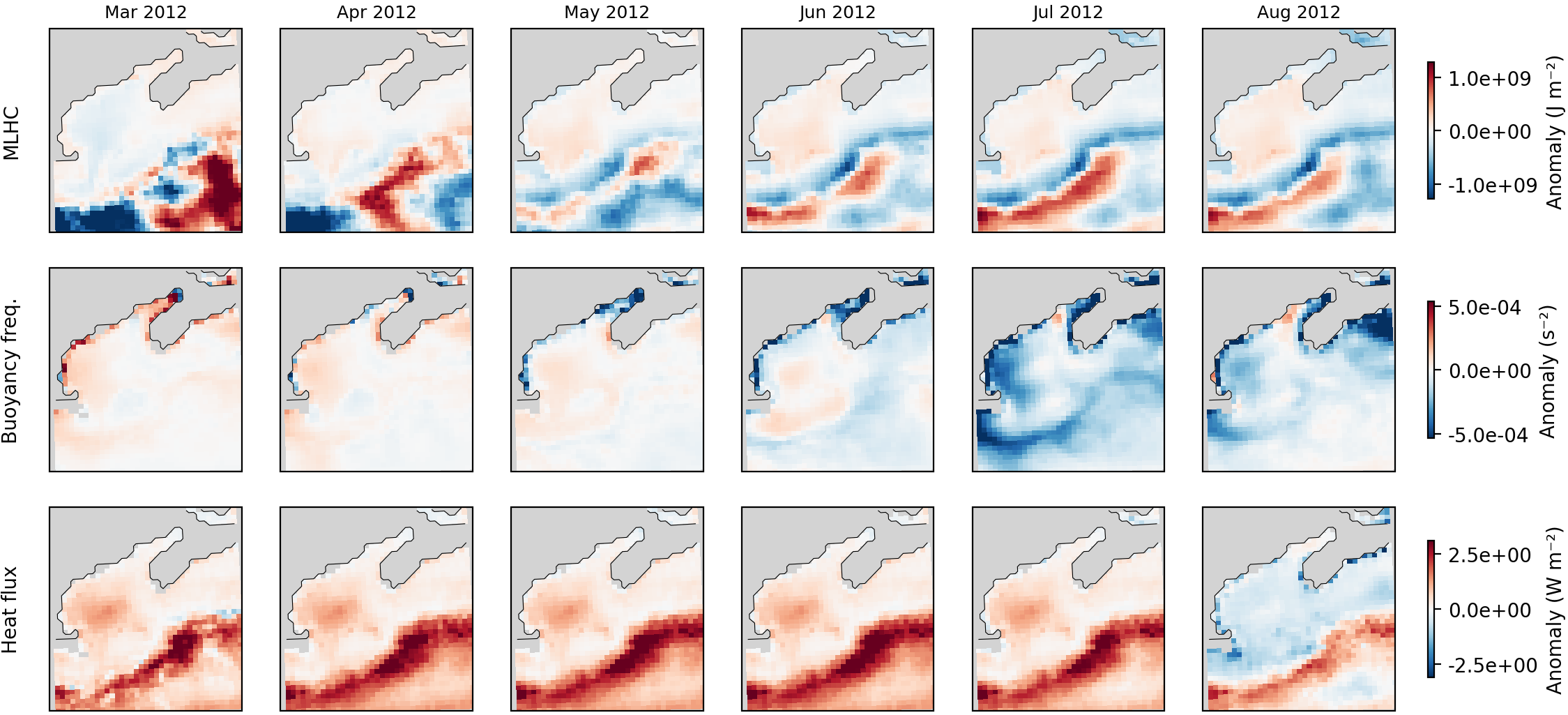}
\caption{\textbf{Six-month retrospective of 2012 Gulf of Maine marine heatwave.}
Six-month retrospective showing anomalies of each field relative to its monthly climatology over 1979--2018.}
\label{fig:6month}
\end{figure}
\paragraph{Mechanistic interpretation of MHW onset.} In Fig. \ref{fig:6month}, we focus on MLHC and the concepts that displayed the most notable anomalous behavior: buoyancy frequency and heat flux entrainment. A small positive anomaly is observed in the MLHC, the first row of Fig. \ref{fig:6month}, indicating that the mixed layer was already warmer than usual in the lead-up to the MHW. We also find a positive anomaly in buoyancy frequency, second row of Fig. \ref{fig:6month}, suggesting enhanced stratification at the base of the mixed layer, which may have inhibited vertical mixing and helped trap heat in the mixed layer for several months. The third row of Fig. \ref{fig:6month} displays that this is accompanied by anomalously high heat flux entrainment, implying persistent atmospheric heat input attempting to penetrate downward into the deeper ocean. However, the strengthened stratification likely reduced the efficiency of this mixing, limiting the redistribution of heat below the mixed layer and allowing warming to accumulate near the surface. Closer to the onset of the MHW, this behavior shifts: stratification anomalies near the surface become negative, indicating that mixing could occur. By August,  heat flux entrainment becomes anomalously low, suggesting a lower temperature gradient at the base of the mixed layer. Despite this possible cooling effect, surface waters may have already accumulated sufficient heat, and anomalous SST conditions manifested, resulting in a MHW.

\paragraph{Agreement with atmospheric findings.} Overall, our analysis suggests MHW development began with anomalous atmospheric heat input that warmed the mixed layer, while increased stratification helped trap that heat near the surface. Although vertical mixing may have contributed to cooling at times, the dominant signal points to atmospheric forcing and heat retention as primary drivers. This is consistent with \cite{chen2014diagnosing} and \cite{chen2015role}, which showed this particular MHW was driven largely by atmospheric heat flux anomalies rather than ocean processes. 

\paragraph{Agreement with ecological findings.} This MHW has well-documented impacts on North American fisheries, as reported by \cite{mills2013fisheries} and \cite{richards2021northern}. The anomalously warm mixed layer conditions predicted by OceanCBM are consistent with these ecological responses, including increased lobster landings associated with warmer waters \cite{mills2013fisheries} and declines in northern shrimp linked to warming-driven ecosystem shifts and predator changes \cite{richards2021northern}. Importantly, these ecological responses cannot be attributed to atmospheric forcing alone, but instead reflect the role of subsurface ocean processes in sustaining and redistributing heat within the mixed layer, as revealed by OceanCBM.

\section{Conclusion}
We introduced an extension of CBMs to disentangle ocean physics. Our approach demonstrates strong predictive skill and interpretability, enabling physically meaningful diagnostics when applied to real-world observed events. Having been rigorously benchmarked against prediction-only and prescription-only baselines, this framework offers relatively consistent mechanistic learning pathways and suggests potential transferability to comparable Earth system models and broader scientific domains involving complex dynamical processes. Limitations and assumptions of this work are discussed in Appendix \ref{s:limits}. 

A key finding is the role of the free concept within the ensemble. Rather than fully predicting the target variable, it captures a meaningful residual signal that is consistent across members but not fully explained by the prescribed concepts, while regularizing weaker concept predictions. This work supports a direction toward mixed supervision modeling in which physical structure is incorporated without overly constraining the learning dynamics. The prescribed concepts gently guide the model toward known physical mechanisms, while the free concept allows flexibility to recover additional structure from data. This balance between prescription and automation provides a grounded pathway for learning interpretable representations of physics.

\bibliographystyle{plain}
\bibliography{references}

@inproceedings{koh2020cbm,
  title={Concept bottleneck models},
  author={Koh, Pang Wei and Nguyen, Thao and Tang, Yew Siang and Mussmann, Stephen and Pierson, Emma and Kim, Been and Liang, Percy},
  booktitle={International conference on machine learning},
  pages={5338--5348},
  year={2020},
  organization={PMLR}
}

@article{van2024timeseries,
  title={Interpretability for Time Series Transformers using A Concept Bottleneck Framework},
  author={van Sprang, Angela and Acar, Erman and Zuidema, Willem},
  journal={arXiv preprint arXiv:2410.06070},
  year={2024}
}

@article{asiyabi2026pgcbm,
  title={Process-Guided Concept Bottleneck Model},
  author={Asiyabi, Reza M and Partnership, SEOSAW and Hancock, Steven and Ryan, Casey},
  journal={arXiv preprint arXiv:2601.10562},
  year={2026}
}

@article{margeloiu2021leakage,
  title={Do concept bottleneck models learn as intended?},
  author={Margeloiu, Andrei and Ashman, Matthew and Bhatt, Umang and Chen, Yanzhi and Jamnik, Mateja and Weller, Adrian},
  journal={arXiv preprint arXiv:2105.04289},
  year={2021}
}

@article{camps2025review,
  title={Artificial intelligence for modeling and understanding extreme weather and climate events},
  author={Camps-Valls, Gustau and Fern{\'a}ndez-Torres, Miguel-{\'A}ngel and Cohrs, Kai-Hendrik and H{\"o}hl, Adrian and Castelletti, Andrea and Pacal, Aytac and Robin, Claire and Martinuzzi, Francesco and Papoutsis, Ioannis and Prapas, Ioannis and others},
  journal={Nature Communications},
  volume={16},
  number={1},
  pages={1919},
  year={2025},
  publisher={Nature Publishing Group UK London}
}

@article{chen2014diagnosing,
  title={Diagnosing the warming of the Northeastern US Coastal Ocean in 2012: A linkage between the atmospheric jet stream variability and ocean response},
  author={Chen, Ke and Gawarkiewicz, Glen G and Lentz, Steven J and Bane, John M},
  journal={Journal of Geophysical Research: Oceans},
  volume={119},
  number={1},
  pages={218--227},
  year={2014},
  publisher={Wiley Online Library}
}

@article{chen2015role,
  title={The role of atmospheric forcing versus ocean advection during the extreme warming of the Northeast US continental shelf in 2012},
  author={Chen, Ke and Gawarkiewicz, Glen and Kwon, Young-Oh and Zhang, Weifeng G},
  journal={Journal of Geophysical Research: Oceans},
  volume={120},
  number={6},
  pages={4324--4339},
  year={2015},
  publisher={Wiley Online Library}
}

@article{richards2021northern,
  title={Northern shrimp Pandalus borealis population collapse linked to climate-driven shifts in predator distribution},
  author={Richards, R Anne and Hunter, Margaret},
  journal={PLoS One},
  volume={16},
  number={7},
  pages={e0253914},
  year={2021},
  publisher={Public Library of Science San Francisco, CA USA}
}

@article{mills2013fisheries,
  title={Fisheries management in a changing climate: lessons from the 2012 ocean heat wave in the Northwest Atlantic},
  author={Mills, Katherine E and Pershing, Andrew J and Brown, Curtis J and Chen, Yong and Chiang, Fu-Sung and Holland, Daniel S and Lehuta, Sigrid and Nye, Janet A and Sun, Jenny C and Thomas, Andrew C and others},
  journal={Oceanography},
  volume={26},
  number={2},
  pages={191--195},
  year={2013},
  publisher={JSTOR}
}

@inproceedings{ronneberger2015u,
  title={U-net: Convolutional networks for biomedical image segmentation},
  author={Ronneberger, Olaf and Fischer, Philipp and Brox, Thomas},
  booktitle={International Conference on Medical image computing and computer-assisted intervention},
  pages={234--241},
  year={2015},
  organization={Springer}
}

@article{murphy1989skill,
  title={Skill scores and correlation coefficients in model verification},
  author={Murphy, Allan H and Epstein, Edward S},
  journal={Monthly weather review},
  volume={117},
  number={3},
  pages={572--582},
  year={1989}
}

@article{zuo2019ecmwf,
  title={The ECMWF operational ensemble reanalysis--analysis system for ocean and sea ice: a description of the system and assessment},
  author={Zuo, Hao and Balmaseda, Magdalena Alonso and Tietsche, Steffen and Mogensen, Kristian and Mayer, Michael},
  journal={Ocean science},
  volume={15},
  number={3},
  pages={779--808},
  year={2019},
  publisher={Copernicus Publications G{\"o}ttingen, Germany}
}

@article{yuksekgonul2022post,
  title={Post-hoc concept bottleneck models},
  author={Yuksekgonul, Mert and Wang, Maggie and Zou, James},
  journal={arXiv preprint arXiv:2205.15480},
  year={2022}
}

@article{sawada2022concept,
  title={Concept bottleneck model with additional unsupervised concepts},
  author={Sawada, Yoshihide and Nakamura, Keigo},
  journal={IEEE Access},
  volume={10},
  pages={41758--41765},
  year={2022},
  publisher={IEEE}
}

@article{oikarinen2023label,
  title={Label-free concept bottleneck models},
  author={Oikarinen, Tuomas and Das, Subhro and Nguyen, Lam M and Weng, Tsui-Wei},
  journal={arXiv preprint arXiv:2304.06129},
  year={2023}
}

@article{suri2026trusting,
  title={Trusting machine learning with physics: A fidelity verification framework for complex systems},
  author={Suri, Sanah and Sonnewald, Maike},
  journal={Authorea Preprints},
  year={2026},
  publisher={Authorea}
}

@article{willard2022integrating,
  title={Integrating scientific knowledge with machine learning for engineering and environmental systems},
  author={Willard, Jared and Jia, Xiaowei and Xu, Shaoming and Steinbach, Michael and Kumar, Vipin},
  journal={ACM Computing Surveys},
  volume={55},
  number={4},
  pages={1--37},
  year={2022},
  publisher={ACM New York, NY}
}

@article{yik2023southern,
  title={Southern ocean dynamics under climate change: New knowledge through physics-guided machine learning},
  author={Yik, William and Sonnewald, Maike and Clare, Mariana CA and Lguensat, Redouane},
  journal={arXiv preprint arXiv:2310.13916},
  year={2023}
}

@article{dheeshjith2025samudra,
  title={Samudra: An AI global ocean emulator for climate},
  author={Dheeshjith, Surya and Subel, Adam and Adcroft, Alistair and Busecke, Julius and Fernandez-Granda, Carlos and Gupta, Shubham and Zanna, Laure},
  journal={Geophysical Research Letters},
  volume={52},
  number={10},
  pages={e2024GL114318},
  year={2025},
  publisher={Wiley Online Library}
}

@article{duncan2025samudrace,
  title={SamudrACE: Fast and accurate coupled climate modeling with 3D ocean and atmosphere emulators},
  author={Duncan, James PC and Wu, Elynn and Dheeshjith, Surya and Subel, Adam and Arcomano, Troy and Clark, Spencer K and Henn, Brian and Kwa, Anna and McGibbon, Jeremy and Perkins, W Andre and others},
  journal={arXiv preprint arXiv:2509.12490},
  year={2025}
}

@article{lam2023learning,
  title={Learning skillful medium-range global weather forecasting},
  author={Lam, Remi and Sanchez-Gonzalez, Alvaro and Willson, Matthew and Wirnsberger, Peter and Fortunato, Meire and Alet, Ferran and Ravuri, Suman and Ewalds, Timo and Eaton-Rosen, Zach and Hu, Weihua and others},
  journal={Science},
  volume={382},
  number={6677},
  pages={1416--1421},
  year={2023},
  publisher={American Association for the Advancement of Science}
}

@article{rudin2022interpretable,
  title={Interpretable machine learning: Fundamental principles and 10 grand challenges},
  author={Rudin, Cynthia and Chen, Chaofan and Chen, Zhi and Huang, Haiyang and Semenova, Lesia and Zhong, Chudi},
  journal={Statistic Surveys},
  volume={16},
  pages={1--85},
  year={2022},
  publisher={The American Statistical Association, the Bernoulli Society, the Institute~…}
}

@article{iglesias2024causally,
  title={Causally-informed deep learning to improve climate models and projections},
  author={Iglesias-Suarez, Fernando and Gentine, Pierre and Solino-Fernandez, Breixo and Beucler, Tom and Pritchard, Michael and Runge, Jakob and Eyring, Veronika},
  journal={Journal of Geophysical Research: Atmospheres},
  volume={129},
  number={4},
  pages={e2023JD039202},
  year={2024},
  publisher={Wiley Online Library}
}

@article{flora2024machine,
  title={A machine learning explainability tutorial for atmospheric sciences},
  author={Flora, Montgomery L and Potvin, Corey K and McGovern, Amy and Handler, Shawn},
  journal={Artificial Intelligence for the Earth Systems},
  volume={3},
  number={1},
  pages={e230018},
  year={2024},
  publisher={American Meteorological Society}
}

@article{cheng2023machine,
  title={Machine learning with data assimilation and uncertainty quantification for dynamical systems: a review},
  author={Cheng, Sibo and Quilodr{\'a}n-Casas, C{\'e}sar and Ouala, Said and Farchi, Alban and Liu, Che and Tandeo, Pierre and Fablet, Ronan and Lucor, Didier and Iooss, Bertrand and Brajard, Julien and others},
  journal={IEEE/CAA Journal of Automatica Sinica},
  volume={10},
  number={6},
  pages={1361--1387},
  year={2023},
  publisher={IEEE}
}

@article{hassija2024interpreting,
  title={Interpreting black-box models: a review on explainable artificial intelligence},
  author={Hassija, Vikas and Chamola, Vinay and Mahapatra, Atmesh and Singal, Abhinandan and Goel, Divyansh and Huang, Kaizhu and Scardapane, Simone and Spinelli, Indro and Mahmud, Mufti and Hussain, Amir},
  journal={Cognitive Computation},
  volume={16},
  number={1},
  pages={45--74},
  year={2024},
  publisher={Springer}
}

@article{karniadakis2021physics,
  title={Physics-informed machine learning},
  author={Karniadakis, George Em and Kevrekidis, Ioannis G and Lu, Lu and Perdikaris, Paris and Wang, Sifan and Yang, Liu},
  journal={Nature Reviews Physics},
  volume={3},
  number={6},
  pages={422--440},
  year={2021},
  publisher={Nature Publishing Group UK London}
}

@article{raissi2019physics,
  title={Physics-informed neural networks: A deep learning framework for solving forward and inverse problems involving nonlinear partial differential equations},
  author={Raissi, Maziar and Perdikaris, Paris and Karniadakis, George E},
  journal={Journal of Computational physics},
  volume={378},
  pages={686--707},
  year={2019},
  publisher={Elsevier}
}

@article{capotondi2024global,
  title={A global overview of marine heatwaves in a changing climate},
  author={Capotondi, Antonietta and Rodrigues, Regina R and Sen Gupta, Alex and Benthuysen, Jessica A and Deser, Clara and Fr{\"o}licher, Thomas L and Lovenduski, Nicole S and Amaya, Dillon J and Le Grix, Natacha and Xu, Tongtong and others},
  journal={Communications Earth \& Environment},
  volume={5},
  number={1},
  pages={701},
  year={2024},
  publisher={Nature Publishing Group UK London}
}

@article{hobday2016hierarchical,
  title={A hierarchical approach to defining marine heatwaves},
  author={Hobday, Alistair J and Alexander, Lisa V and Perkins, Sarah E and Smale, Dan A and Straub, Sandra C and Oliver, Eric CJ and Benthuysen, Jessica A and Burrows, Michael T and Donat, Markus G and Feng, Ming and others},
  journal={Progress in oceanography},
  volume={141},
  pages={227--238},
  year={2016},
  publisher={Elsevier}
}

@article{spillman2025makes,
  title={What makes a marine heatwave forecast useable, useful and used?},
  author={Spillman, Claire M and Hobday, Alistair J and Behrens, Erik and Feng, Ming and Capotondi, Antonietta and Cravatte, Sophie and Holbrook, Neil J and Gupta, Alex Sen},
  journal={Progress in Oceanography},
  volume={234},
  pages={103464},
  year={2025},
  publisher={Elsevier}
}

@article{sun2023frequent,
  title={Frequent marine heatwaves hidden below the surface of the global ocean},
  author={Sun, Di and Li, Furong and Jing, Zhao and Hu, Shijian and Zhang, Bohai},
  journal={Nature Geoscience},
  volume={16},
  number={12},
  pages={1099--1104},
  year={2023},
  publisher={Nature Publishing Group UK London}
}

@article{ren2026ocean,
  title={Ocean dynamics shape marine heatwaves and their predictability},
  author={Ren, Xianglin and Liu, Wei and Zhang, Liping},
  journal={Nature Communications},
  year={2026},
  publisher={Nature Publishing Group UK London}
}

@article{singh2024rethinking,
  title={Rethinking interpretability in the era of large language models},
  author={Singh, Chandan and Inala, Jeevana Priya and Galley, Michel and Caruana, Rich and Gao, Jianfeng},
  journal={arXiv preprint arXiv:2402.01761},
  year={2024}
}

@article{vijith2020closing,
  title={Closing the sea surface mixed layer temperature budget from in situ observations alone: Operation Advection during BoBBLE},
  author={Vijith, V and Vinayachandran, PN and Webber, Benjamin GM and Matthews, Adrian J and George, Jenson V and Kannaujia, Vijay Kumar and Lotliker, Aneesh A and Amol, P},
  journal={Scientific reports},
  volume={10},
  number={1},
  pages={7062},
  year={2020},
  publisher={Nature Publishing Group UK London}
}

@article{hersbach2020era5,
  title={The ERA5 global reanalysis},
  author={Hersbach, Hans and Bell, Bill and Berrisford, Paul and Hirahara, Shoji and Hor{\'a}nyi, Andr{\'a}s and Mu{\~n}oz-Sabater, Joaqu{\'\i}n and Nicolas, Julien and Peubey, Carole and Radu, Raluca and Schepers, Dinand and others},
  journal={Quarterly journal of the royal meteorological society},
  volume={146},
  number={730},
  pages={1999--2049},
  year={2020},
  publisher={Wiley Online Library}
}

@article{lang2024aifs,
  title={AIFS--ECMWF's data-driven forecasting system},
  author={Lang, Simon and Alexe, Mihai and Chantry, Matthew and Dramsch, Jesper and Pinault, Florian and Raoult, Baudouin and Clare, Mariana CA and Lessig, Christian and Maier-Gerber, Michael and Magnusson, Linus and others},
  journal={arXiv preprint arXiv:2406.01465},
  year={2024}
}

@article{sonnewald2021bridging,
  title={Bridging observations, theory and numerical simulation of the ocean using machine learning},
  author={Sonnewald, Maike and Lguensat, Redouane and Jones, Daniel C and Dueben, Peter D and Brajard, Julien and Balaji, Venkatramani},
  journal={Environmental Research Letters},
  volume={16},
  number={7},
  pages={073008},
  year={2021},
  publisher={IOP Publishing}
}

@article{sonnewald2021revealing,
  title={Revealing the impact of global heating on North Atlantic circulation using transparent machine learning},
  author={Sonnewald, Maike and Lguensat, Redouane},
  journal={Journal of Advances in Modeling Earth Systems},
  volume={13},
  number={8},
  pages={e2021MS002496},
  year={2021},
  publisher={Wiley Online Library}
}

@article{irrgang2021towards,
  title={Towards neural Earth system modelling by integrating artificial intelligence in Earth system science},
  author={Irrgang, Christopher and Boers, Niklas and Sonnewald, Maike and Barnes, Elizabeth A and Kadow, Christopher and Staneva, Joanna and Saynisch-Wagner, Jan},
  journal={Nature Machine Intelligence},
  volume={3},
  number={8},
  pages={667--674},
  year={2021},
  publisher={Nature Publishing Group UK London}
}

@article{guo2025data,
  title={Data-driven global ocean modeling for seasonal to decadal prediction},
  author={Guo, Zijie and Lyu, Pumeng and Ling, Fenghua and Bai, Lei and Luo, Jing-Jia and Boers, Niklas and Yamagata, Toshio and Izumo, Takeshi and Cravatte, Sophie and Capotondi, Antonietta and others},
  journal={Science Advances},
  volume={11},
  number={33},
  pages={eadu2488},
  year={2025},
  publisher={American Association for the Advancement of Science}
}

@article{pasula2025data,
  title={Data Driven Deep Learning for Correcting Global Climate Model Projections of SST and DSL in the Bay of Bengal},
  author={Pasula, Abhishek and Subramani, Deepak N},
  journal={arXiv preprint arXiv:2504.20620},
  year={2025}
}

@misc{SDSCExpanse,
  author = {{San Diego Supercomputer Center}},
  title  = {Expanse},
  year   = {2025},
  howpublished = {University of California San Diego},
  doi    = {10.1145/3437359.3465588},
  url    = {https://doi.org/10.1145/3437359.3465588}
}

@article{krishnapriyan2021characterizing,
  title={Characterizing possible failure modes in physics-informed neural networks},
  author={Krishnapriyan, Aditi and Gholami, Amir and Zhe, Shandian and Kirby, Robert and Mahoney, Michael W},
  journal={Advances in neural information processing systems},
  volume={34},
  pages={26548--26560},
  year={2021}
}

@article{radfar2026synergistic,
  title={Synergistic impact of marine heat waves and rapid intensification exacerbates tropical cyclone destructive power worldwide},
  author={Radfar, Soheil and Foroumandi, Ehsan and Moftakhari, Hamed and Moradkhani, Hamid and Gupta, Alex Sen and Foltz, Gregory R},
  journal={Science Advances},
  volume={12},
  number={15},
  pages={eadu1733},
  year={2026},
  publisher={American Association for the Advancement of Science}
}

@article{mccabe2016unprecedented,
  title={An unprecedented coastwide toxic algal bloom linked to anomalous ocean conditions},
  author={McCabe, Ryan M and Hickey, Barbara M and Kudela, Raphael M and Lefebvre, Kathi A and Adams, Nicolaus G and Bill, Brian D and Gulland, Frances MD and Thomson, Richard E and Cochlan, William P and Trainer, Vera L},
  journal={Geophysical research letters},
  volume={43},
  number={19},
  pages={10--366},
  year={2016},
  publisher={Wiley Online Library}
}

\clearpage

\appendix 

\section{Limitations}\label{s:limits}
While our results are promising, there are certain limitations and assumptions we acknowledge. First, we performed the OceanCBM, prescription-only and prediction-only model configurations using the same random seed for the individual members. Specifically, v1 for OceanCBM uses the same random seed as v1 for prescription-only and prediction-only configurations, and so on. While this enables direct comparison of mechanistic learning pathways across configurations, it operates on the assumption that random initialization alone captures sufficient variation in learned representations. Second, the prescribed concept set is necessarily non-exhaustive, reflecting cascading limitations from oceanography: the drivers of MLHC are not fully established and the system's complexity means nonlinear relationships may exist in the data that are not fully realized through the prescribed concepts. Finally, while the free concept's simplicity and dual role is itself an interesting result, its lack of constraints leaves its exact physical interpretation as an open question for future work.

\section{Broader impacts}\label{s:impacts}
Marine heatwaves (MHWs) are increasing in frequency and intensity, posing growing risks to coastal ecosystems, fisheries, and regional economies. They have been known to cause wildlife mortality events, coral bleaching and even `supercharge' tropical cyclones \cite{capotondi2024global, radfar2026synergistic}. There are often precursors to anomalously warm sea surface temperatures that are not considered rigorously in current MHW forecasting, limiting our ability to understand the mechanisms underlying MHW onset and evolution. By explicitly modeling these intermediate physical quantities, OceanCBM provides a means to interrogate forecast systems in terms of physically meaningful mechanisms rather than surface-level anomalies alone. For example, if the model detects elevated MLHC alongside increased stratification one month in advance, and an operational forecast from NOAA or ECMWF indicates a potential heatwave in the same region at a longer lead time, this earlier signal of excess ocean heat could inform fisheries managers to implement mitigation strategies such as early harvesting. Similarly, when MLHC reaches levels that may contribute to harmful algal blooms, identifying these conditions on subseasonal timescales provides valuable lead time for public health officials to prepare for potential risks associated with seafood pathogens \cite{mccabe2016unprecedented}. By linking prediction with mechanism, OceanCBM unlocks more informed and timely responses to emerging ocean conditions. This aligns with the broader requirements for actionable climate risk assessment, which emphasize not only predictive skill but also interpretability, usability, and effective communication between model developers and end users \cite{spillman2025makes}. In this sense, OceanCBM complements existing forecasting systems by enhancing their interpretability and enabling deeper scientific understanding of extreme ocean events. We note however that OceanCBM is trained on monthly-averaged reanalysis data over the North Atlantic region. While promising, generalization to other ocean basins or finer temporal scales should be approached keeping this in mind.

\section{Data}\label{s:data}
All data used in this work are derived from the ensemble members of ORAS5 \cite{zuo2019ecmwf}. In Tables \ref{tab:input} and \ref{tab:concepts}, bold labels indicate variables derived in this work, while non-bold labels correspond to quantities taken directly from ORAS5. Table \ref{tab:input} describes the inputs, while Table \ref{tab:concepts} describes the concepts.
\newpage
\begin{table}[h]
\begin{tabularx}{\textwidth}{@{} >{\raggedright\arraybackslash}p{3.5cm} l X @{}}
\toprule
\textbf{Input} & \textbf{Label} & \textbf{Description} \\
\midrule
Sea surface temperature [°C] & \texttt{sosstsst} & Temperature of the ocean surface. \\
\addlinespace
Sea surface salinity [psu] & \texttt{sosaline} & Salt concentration of the ocean surface. \\
\addlinespace
Sea surface height [m] & \texttt{sossheig} & Deviation of sea level from its mean state at a given location. \\
\addlinespace
Mixed layer depth (MLD) [m] & \texttt{somxl010} & Depth at which potential density increases by 0.01 kg m$^{-3}$ relative to a surface reference. \\
\addlinespace
Net downward heat flux [W m$^{-2}$] & \texttt{sohefldo} & Net energy flux entering the ocean at the surface. \\
\addlinespace
Zonal velocity [m s$^{-1}$] & \textbf{\texttt{vozocrtx\_ml}} & East–west velocity averaged across the mixed layer. \\
\addlinespace
Meridional velocity [m s$^{-1}$] & \textbf{\texttt{vomecrty\_ml}} & North–south velocity averaged across the mixed layer. \\
\addlinespace
Potential temperature difference at MLD [°C] & \textbf{\texttt{votempdiff}} & Difference in potential temperature across the mixed layer base. \\
\addlinespace
Salinity difference at MLD [psu] & \textbf{\texttt{vosaldiff}} & Difference in salinity across the mixed layer base. \\
\addlinespace
Bathymetry & \texttt{mbathy} & Ocean floor topography. \\
\addlinespace
Coriolis parameter &  \texttt{ff} & Parameter capturing the effect of Earth’s rotation on moving fluids. \\
\addlinespace
Wind stress curl & \textbf{\texttt{sowsc}} & Measure of wind-driven rotational forcing on the ocean surface. \\
\bottomrule
\end{tabularx}
\vspace{1.5ex}
\caption{Descriptions of input variables.}
\label{tab:input}
\end{table}
\begin{table}[!h]
\begin{tabularx}{\textwidth}{@{} >{\raggedright\arraybackslash}p{3.5cm} l X @{}}
\toprule
\textbf{Concept} & \textbf{Label} & \textbf{Description} \\
\midrule
Vertical shear & \textbf{\texttt{vos2}} & How the horizontal movement of water changes with depth, calculated as $S^2 = |\frac{\partial u}{\partial z}|^2 + |\frac{\partial v}{\partial z}|^2$ using zonal and meridional velocity at the MLD base. Indicative of turbulent entrainment. \\
\addlinespace
Buoyancy frequency & \textbf{\texttt{von2}} & Measures the density gradient or resistance to vertical mixing, defined as $N^2 = \frac{g}{\rho_0}\frac{\partial \rho}{\partial z}$ using potential temperature and salinity at the MLD. \\
\addlinespace
Heat flux entrainment & \textbf{\texttt{vohfe}} & The heat exchange occurring during mixed layer deepening: $Q_e = \rho \cdot c_p \cdot w_e \cdot \Delta T$, where $w_e = \max(\frac{\partial h}{\partial t}, 0)$ and $\Delta T$ is the temperature difference across the MLD base. \\
\addlinespace
Mixed layer depth tendency & \textbf{\texttt{mxl\_tendency}} & The rate at which the mixed layer depth changes over time, represented as $\frac{\partial z}{\partial t}$. \\
\bottomrule
\end{tabularx}
\vspace{1.5ex}
\caption{Description of derived concepts.}
\label{tab:concepts}
\end{table}

\textbf{Data Preprocessing}
\begin{itemize}
    \item \textbf{Detrend:} All the input data and concepts (after calculation) are detrended. We fit a linear trend to the data over the entire evaluation period of $40$ years, and subtract this trend per grid cell. 
    \item \textbf{Clipping}: To facilitate heavy outliers, we clip at the $2^{\text{nd}}$ and $98^{\text{th}}$ percentile. 
    \item \textbf{Smoothing}: A Gaussian filter with $\sigma=3$ is applied to data with grid artifacts from concept calculations. 
\end{itemize}

\section{Training}\label{s:training}
\begin{itemize}
    \item \textbf{Optimizer:} All models were trained using the AdamW optimizer with a learning rate of $0.001$ and a weighted decay of $0.0001$ for $100$ epochs and a batch size of $64$.
    \item \textbf{Loss Schedule:} The loss is a weighted combination of concept and prediction loss. $$\mathcal{L}_{\text{combined}} = \lambda \mathcal{L}_{\text{concept}} + (1-\lambda)\mathcal{L}_{\text{pred}}$$ We train with $\lambda = 0$ for the prediction-only model. For OceanCBM, $\lambda$ starts at $0.8$ and exponentially decays to $0.2$.
    \item \textbf{Ensembling:} OceanCBM is an ensemble of $5$ members that all have independent random weight initialization. The ensemble members of prediction-only and prescription-only configurations are trained with the same random weight initializations as their OceanCBM counterparts. OceanCBM's concept and target predictions are the result of averages across ensemble members. 
    \item \textbf{Normalization:} The inputs and concepts are normalized using Z-score normalization. The normalization statistics are taken across the training set and applied to the training, testing and validation sets. 
    \item \textbf{Hardware:} Our models were trained and tested on SDSC's Expanse using 1 NVIDIA V100 \cite{SDSCExpanse}. Training spanned approximately $85$ minutes per ensemble member.
\end{itemize}

\section{Extended Results}
\begin{figure}[htbp]
    \centering
    % First Subfigure: Vertical Shear
    \begin{subfigure}[b]{0.48\textwidth}
        \centering
        \includegraphics[width=\linewidth]{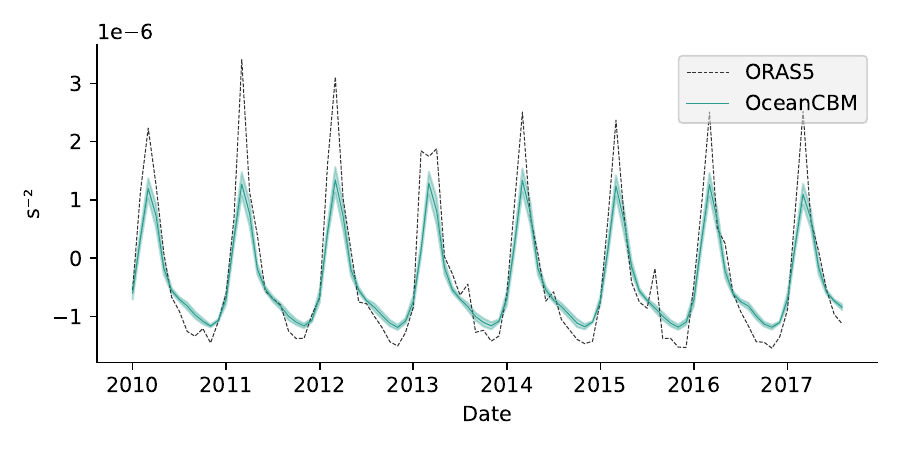}
        \caption{Vertical shear construction.}
        \label{fig:vos2_ts}
    \end{subfigure}
    \hfill % This adds horizontal space between the two images
    % Second Subfigure: MLD Tendency
    \begin{subfigure}[b]{0.48\textwidth}
        \centering
        \includegraphics[width=\linewidth]{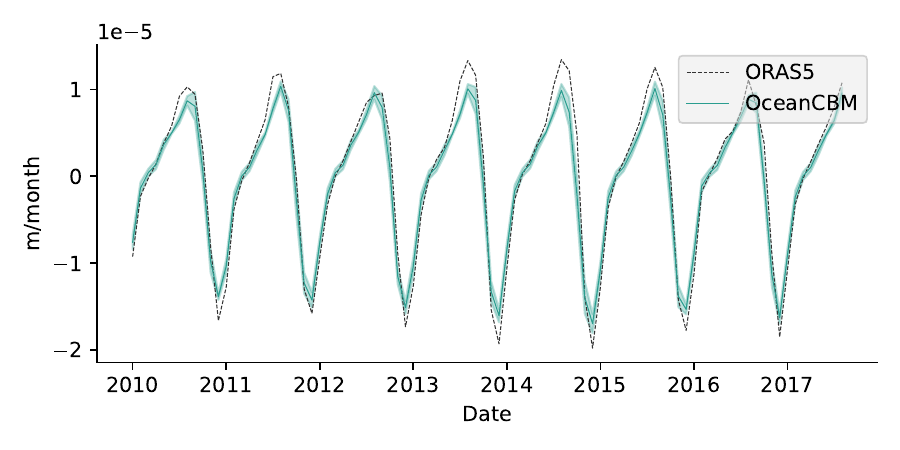}
        \caption{MLD tendency construction.}
        \label{fig:mxl_tendency_ts}
    \end{subfigure}
    
    \caption{Out-of-sample time-series reconstructions for vertical shear and mixed layer depth tendency. The shaded region denotes $\pm 2$ standard deviations across the ensemble members, to visualize model spread.}
    \label{fig:ts_constructions}
\end{figure}

In this section, we detail the anomaly correlation coefficients (ACC) for OceanCBM, the prediction-only model, and the prescription-only model. We additionally provide the time-series plots (extensions of Fig. \ref{fig:ts_constructions}) for mixed layer depth tendency and vertical shear. And finally, we provide all the seasonal discrepancies between the discovered free concept and predicted mixed layer heat content in the validation and test set. 
\begin{table}[h]
    \centering
    \label{tab:seasonal_acc}
    \small % Adjusting font size to fit width if necessary
    \begin{tabular}{llcccc}
        \toprule
        \textbf{Variable} & \textbf{Config} & \textbf{DJF} & \textbf{MAM} & \textbf{JJA} & \textbf{SON} \\
        \midrule
        \multirow{3}{*}{MLHC} & OceanCBM      & 0.9116 & 0.9174 & 0.9191 & 0.9207 \\
        & Prediction-only & 0.9131 & 0.9186 & 0.9180 & 0.9216 \\
        & Prescription-only      & 0.9129 & 0.9187 & 0.9188 & 0.9212 \\
        \midrule
        \multirow{2}{*}{MLD tendency} & OceanCBM & 0.8384 & 0.8127 & 0.8302 & 0.7968 \\
        & Prescription-only & 0.8202 & 0.7871 & 0.8108 & 0.7790 \\
        \midrule
        \multirow{2}{*}{Buoyancy frequency} & OceanCBM & 0.8014 & 0.8159 & 0.8072 & 0.8103 \\
        & Prescription-only & 0.8069 & 0.8204 & 0.8107 & 0.8146 \\
        \midrule
        \multirow{2}{*}{Heat flux entrainment} & OceanCBM & 0.7532 & 0.7413 & 0.7700 & 0.7599 \\
        & Prescription-only & 0.7397 & 0.7102 & 0.7505 & 0.7329 \\
        \midrule
        \multirow{2}{*}{Vertical shear} & OceanCBM & 0.5965 & 0.6145 & 0.6727 & 0.5818 \\
        & Prescription-only & 0.5332 & 0.5709 & 0.5774 & 0.5755 \\
        \bottomrule
    \end{tabular}
    \vspace{1.5ex}
    \caption{Seasonal Anomaly Correlation Coefficient (ACC) across Model Configurations.}
\end{table}

\begin{figure}[h]
    \centering
    \includegraphics[width=\linewidth]{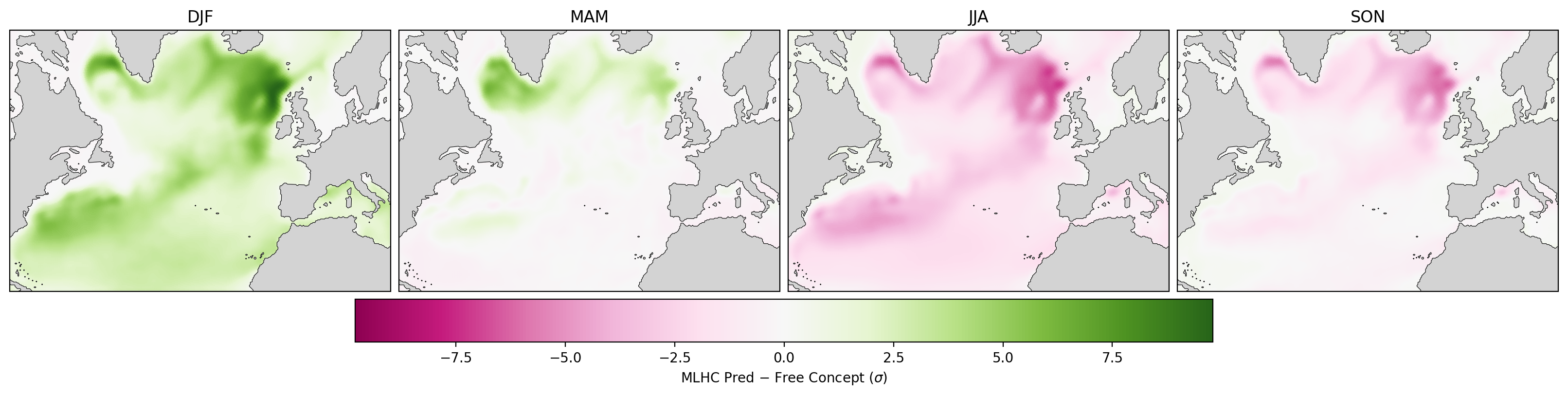}
    \caption{Seasonal discrepancies between the predicted mixed layer heat content and the learned free concept. The discrepancies are most pronounced in peak summer and winter months.}
    \label{fig:placeholder}
\end{figure}

\newpage

\end{document}